# Aviation Safety Enhancement via NLP & Deep Learning: Classifying Flight Phases in ATSB Safety Reports

Aziida Nanyonga
School of Engineering and Information Technology
University of New South Wales
Canberra, Australia
a.nanyonga@adfa.edu.au

Hassan Wasswa
School of Engineering and Information Technology
University of New South Wales
Canberra, Australia
h.wasswa@adfa.edu.au

Graham Wild
School of Engineering and Information Technology
University of New South Wales
Canberra, Australia
g.wild@adfa.edu.au

*Abstract*— Aviation safety is paramount, demanding precise analysis of safety occurrences during different flight phases. This study employs Natural Language Processing (NLP) and Deep Learning models, including LSTM, CNN, Bidirectional LSTM (BLSTM), and simple Recurrent Neural Networks (sRNN), to classify flight phases in safety reports from the Australian Transport Safety Bureau (ATSB). The models exhibited high accuracy, precision, recall, and F1 scores, with LSTM achieving the highest performance of 87%, 88%, 87%, and 88%, respectively. This performance highlights their effectiveness in automating safety occurrence analysis. The integration of NLP and Deep Learning technologies promises transformative enhancements in aviation safety analysis, enabling targeted safety measures and streamlined report handling.

## I. INTRODUCTION

The aviation industry is one of the most regulated and safety-conscious sectors globally, driven by a commitment to ensuring passenger safety and operational excellence. To maintain and improve aviation safety, it is crucial to investigate and analyze safety occurrences systematically [1]. These occurrences range from technical malfunctions to human errors, and they occur throughout different phases of flight, including pre-flight, take-off, climb, cruise, descent, approach, and landing. Understanding when and where these occurrences are more likely to happen is vital for targeted safety measures [2]. Historically, safety occurrence reports have been a valuable resource for aviation safety analysis. They provide detailed narratives of incidents, accidents, and near-miss events, offering valuable context and information about what transpired during these events. However, analyzing these reports manually is a resource-intensive task, often limited by human capacity, subjectivity, and the potential for oversight [3].

The motivation behind this research stems from the need to automate and streamline the analysis of safety occurrence reports, specifically in the context of flight phase classification. Effective classification of safety occurrences into their corresponding flight phases can lead to a more precise understanding of when and where these incidents are likely to occur. This knowledge, in turn, enables the development of targeted safety protocols and preventative measures to reduce the frequency and severity of incidents. Moreover, by automating this process, aviation safety authorities can handle a higher volume of reports efficiently, ensuring that critical information is not overlooked, and safety enhancements are implemented promptly [4]. This research harnesses the potential of NLP and Deep Learning to achieve this goal, promising to revolutionize aviation safety analysis [5].

The primary objective of this study is to investigate the effectiveness of NLP and Deep Learning techniques in classifying flight phases within safety occurrence reports obtained from the ATSB. To achieve this, we employed advanced deep learning architectures, including LSTM, CNN, BLSTM, and sRNN models [6], [7]. These models are trained to infer flight phase information from the unstructured text narratives in the safety occurrence reports. Evaluation metrics including accuracy, precision, recall, and F1-score were used for measuring model performance. The goal is to demonstrate that NLP and Deep Learning models can effectively infer flight phase information from raw text narratives, providing a foundation for more comprehensive safety occurrence analysis.

The structure of this paper is as follows: Section II provides a review of the existing literature, highlighting the significance of flight phase classification in aviation safety research and discussing relevant prior work. Section III gives an account of the methodology employed in this study, including data preprocessing, model selection, training, and evaluation. Section IV presents the results of our experiments, showcasing the performance of the deep learning models in flight phase classification and discussing the interpretation of results, potential limitations, and implications for aviation safety. Finally, Section V concludes the paper by summarizing the key findings and their significance in enhancing aviation safety.

## II. RELATED WORK

The classification of flight phases within safety occurrence reports using NLP and Deep Learning techniques represents a significant advancement in aviation safety analysis [7]–[11]. This section explores prior work related to flight phase classification, emphasizing the growing importance of automated methods.

Understanding the significance of flight phase classification within aviation safety analysis is critical. Researchers have long recognized the importance of associating safety occurrences with specific flight phases to target preventive measures effectively [12], [13].

Historically, the classification of safety occurrences into flight phases has been a labor-intensive, manual process. Aviation safety experts and investigators typically reviewed incident narratives and categorized them based on their expertise and experience. This approach, while valuable, is limited by subjectivity and resource constraints [3]. Comparative studies have been conducted to evaluate the performance of various classification methods. Study [14] compared traditional rule-based classification with machine learning approaches, finding that machine learning outperformed rule-based methods in accuracy and efficiency. However, the study did not explore the full potential of Deep Learning techniques.

The International Civil Aviation Organization (ICAO) introduced a taxonomy system for flight phase classification, providing a standardized framework for manual categorization. While this system has been widely adopted, its effectiveness depends on the availability of expert human resources [15]. The emergence of NLP and Deep Learning has revolutionized the way safety occurrences are classified by enabling automated and scalable solutions. Researchers have explored various approaches to automate flight phase classification [16].

In research [17] a study was conducted on aviation safety and emphasized the need for precise classification of incidents by flight phases based on visual scanning strategies using SVM. The study found that incidents occurring during take-off and landing phases tend to have different causation factors than those during cruising, highlighting the importance of context-aware analysis.

Also, [16] employed two deep learning techniques, ResNet and sRNN, to classify flight phases based on textual data extracted from the NTSB dataset. Impressively, their models achieved an accuracy exceeding 68%, significantly surpassing the random guess rate of 14% for a seven-class classification problem. Moreover, these models demonstrated exceptional precision, recall, and F1 scores, with sRNN notably outperforming the simplified ResNet architecture.

Our study extends the existing work in several ways. Firstly, we explore the use of advanced Deep Learning models in flight phase classification, including LSTM, CNN, BLSTM, and sRNN [7], which have not been extensively studied in this context. Secondly, we employ a substantial dataset of 53,275 safety occurrence reports from the ATSB to evaluate the performance of these models rigorously. Thirdly, we focus on comprehensive evaluation metrics, including accuracy, precision, recall, and F1-score, to provide a holistic assessment of the models' performance.

## III. METHODOLOGY

In this section, we outline the methodology employed in this research to classify flight phases within safety occurrence reports from the ATSB using NLP and Deep Learning techniques. Our approach involves data preprocessing, model selection, training, and evaluation as depicted in Fig. 1.

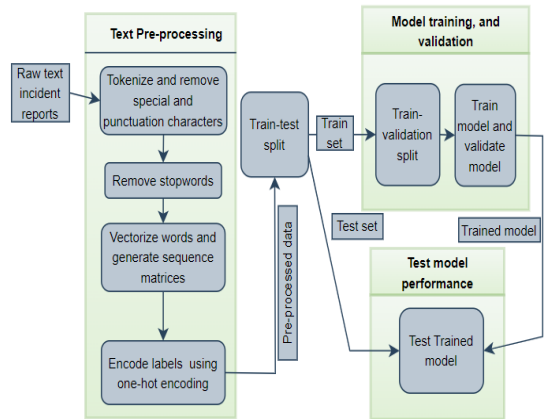

Fig. 1. Methodological framework

### A. Data Acquisition

Aviation incident/accident investigation reports are collected and published by various organizations such as ATSB, the Aviation Safety Reporting System (ASRS), and the National Transportation Safety Board (NTSB). For this study, the researchers utilized the ATSB aviation incident/accident investigation reports. Depending on the nature of the problem, we considered text narratives and phase(s) of flight that were recorded in Australia for the period of 10 years resulting in a dataset with 53,275 records where the data was sourced directly from the ATSB investigation authorities spanning from 1/01/2013 to 12/31/2022. Moreover, the researchers focused on phase (s) of flight, resulting in a dataset comprising 50,778 records following data preprocessing and cleaning. From each report, the 'Summary' and 'phase (s) of flight' fields were extracted for training and validation of deep learning models.

### B. Text Processing

Text preprocessing plays a crucial role in preparing unstructured text data for machine learning models. In this study, we leveraged the Keras deep learning library for its extensive collection of deep learning models and model layers. Additionally, Keras provides advanced modules for text preprocessing. Specifically, we utilized the Tokenizer module, which efficiently generates tokens and sequence vectors from input text. To encode categorical data, such as phase (s) of flight labels (e.g., take-off, initial climb, approach, landing, manoeuvring, cruise, descent, standing, taxiing, and others), we employed the tocategorical module in Keras, mapping these categorical entries to numerical values using one-hot encoding for each data instance.

To address challenges related to special characters, punctuation, and stop words, as well as to perform word lemmatization, we harnessed the capabilities of the spacy library. Spacy is a Python library tailored for text-processing tasks, encompassing functionalities like named entity recognition and word tagging. It maintains an extensive list of special characters, punctuation marks, and stop words and undergoes regular updates whenever necessary to remain current.

With the aforementioned tools at our disposal, each input text narrative underwent a comprehensive preprocessing pipeline, ultimately being transformed into a representative sequence or vector with a fixed length of 2000. For narratives with fewer than 2000 words, we padded the numeric sequences with zeros, ensuring uniformity. In contrast, narratives exceeding 2000 words were truncated to meet this standardized length. The vocabulary size of the text corpus was set to 100,000, accommodating a broad range of terms.

To partition the dataset into training (80%), and testing (20%) sets, we utilized the train-test-split module from scikit-learn. All experiments conducted in this study were implemented using the Python programming language, with Jupyter Notebook serving as the chosen code editor. This rigorous text preprocessing framework laid the foundation for subsequent model training and evaluation, enabling the accurate classification of phase (s) of flight based on unstructured safety occurrence narratives from the ATSB dataset.

*C. Text Classification*

To ensure model robustness and prevent overfitting during the training, 10% of the train-set was set aside for model validation in each training epoch. This practice facilitated continuous evaluation and refinement of the models. Four distinct deep learning architectures, namely LSTM, BLSTM, CNN, and sRNN, were trained on this data, each offering unique capabilities for text classification tasks. Model optimization was accomplished using the Adam optimizer, chosen for its efficiency in gradient-based optimization. It is noteworthy that this study did not focus explicitly on identifying the best optimizer, thus allowing for the exploration of alternative optimization techniques in future research endeavours.

*D. Deep Learning Model Architecture*

For consistency and comparability across all models, a shared architecture served as the foundation, with slight adjustments for each model. This standardized architecture comprised three key components: an embedding layer, hidden layers, and an output layer. To introduce non-linearity and capture complex relationships in the data, the Rectified Linear Unit (ReLU) activation function was applied to all hidden layers. Meanwhile, the SoftMax activation function was adopted for the output layer, facilitating multi-class classification. The final predicted class was determined using the argmax function, which identifies the index associated with the highest probability in the SoftMax output. For a visual representation of the deep learning architectures employed in this study, please refer to Fig. 2.

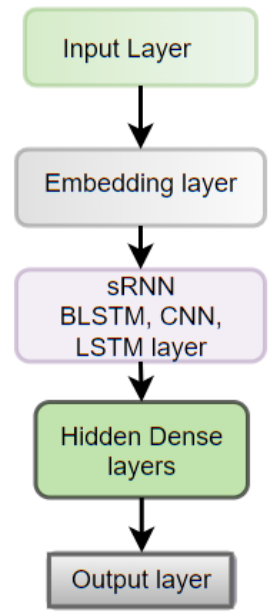

Fig. 2. Deep learning architectures

This consistent architecture provided a solid foundation for training and evaluation of the deep learning models, enabling a fair comparison of their performance and the accurate classification of flight phases based on unstructured safety occurrence narratives.

*E. Model Performance Evaluation*

This section elucidates the evaluation criteria utilized in this study to assess the models' performance. The primary focus of this research is multi-class classification, and as such, performance was gauged based on the accuracy of predictions across various classes. To comprehensively evaluate model performance, we employed a suite of standard prediction performance metrics, including precision, recall, F1-score, and accuracy, as cited in prior literature [18], [19]. These metrics, explained in Table 1, provide a comprehensive assessment of the models' classification performance.

TABLE I. SUMMARY OF EVALUATION METRICS

| Metrics | Formula | Evaluation focus |
|---|---|---|
| Precision (p) | $\frac{TP}{TP + FP}$ | Correctly predicted positives in a positive class |
| Recall (r) | $\frac{TP}{TP + TN}$ | Fraction of positive patterns correctly classified |
| F1-score (F) | $\frac{2 * precision * recaal}{precision + recall}$ | Weighted average score of precision and recall |
| Accuracy (acc) | $\frac{TP + TN}{TP + FP + TN + FN}$ | Total number of instances predicted correctly |

*1) Confusion matrix*

A confusion matrix is an invaluable tool for visually assessing model performance in a classification task. As illustrated in Table II, it is a square matrix in the dimensional space, m, where m is the number of unique entries in the dependent variable, essentially reflecting instances distributed among class labels during the testing phase of the AI model. The confusion matrix provides a clear visualization of the models' performance, serving as a yardstick to gauge their effectiveness. In the matrix shown in Table II, diagonal cells denote correct predictions, including true positives (TP) and true negatives (TN), while off-diagonal cells indicate incorrect predictions, encompassing false negatives (FN) and false positives (FP) [20].

TABLE II. CONFUSION MATRIX

| Actual Value | Predicted Value | |
|---|---|---|
| | TN | FP |
| | FN | TP |

This robust evaluation approach enabled us to comprehensively assess the models' classification accuracy, precision, recall, and F1 score, providing a holistic understanding of its performance in classifying flight phases. [7].

## IV. RESULTS AND DISCUSSION

In this study, we have explored the application of NLP and Deep Learning techniques to enhance aviation safety analysis by classifying flight phases within safety occurrence reports. Leveraging a substantial dataset of 50,778 safety occurrence reports provided by ATSB, we evaluated the performance of advanced Deep Learning models, including LSTM, CNN, BLSTM, and sRNN, using a variety of performance metrics.

### A. Model Performance

Our findings reveal the remarkable capabilities of these models in accurately classifying flight phases within unstructured text narratives. Table III shows the key performance results for each model.

TABLE III. DEEP LEARNING MODEL PERFORMANCE

| Models | Precision (%) | Recall (%) | F1 (%) | Accuracy (%) |
|---|---|---|---|---|
| **LSTM** | **88** | 87 | **88** | **87.4** |
| **sRNN** | 78 | 77 | 77 | 77.0 |
| **BLSTM** | **88** | 87 | 87 | **87.3** |
| **CNN** | 87 | 87 | 87 | 86.5 |

These results clearly demonstrate the effectiveness of NLP and Deep Learning models in handling the complexity of aviation safety reports and inferring crucial flight phase information. Notably, the LSTM and BLSTM models exhibited the highest accuracy and precision, indicating their suitability for this task.

Both Fig. 3 and Fig. 4, illustrate the training behavior and performance of the four deep learning models (SRNN, BLTM, CNN, and LSTM) in terms of validation accuracy and validation loss, respectively, over a range of training epochs. These visualizations provide valuable information for assessing and comparing the models' capabilities in solving the classification problem at hand.

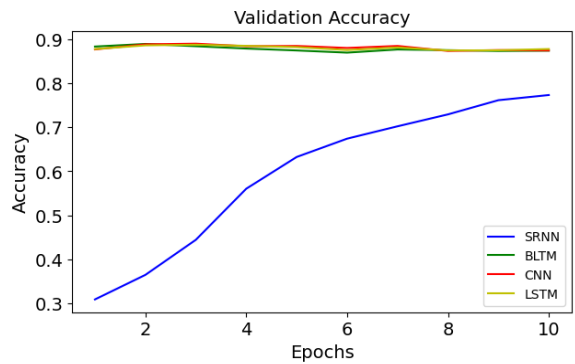

Fig. 3. Validation accuracy performance

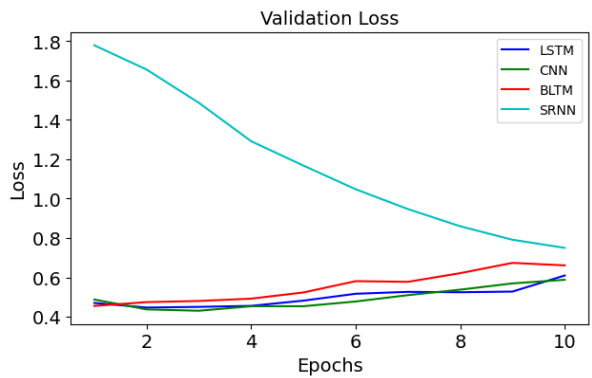

Fig. 4. Validation Loss

```
Classification Report
------------------------------------------------------
                    precision    recall  f1-score   support

           Approach      0.86      0.91      0.88      1915
              Climb      0.86      0.91      0.88      1253
             Cruise      0.83      0.88      0.86      1267
            Landing      0.97      0.87      0.92      1908
Manoeuvering/airwork      0.84      0.84      0.84      1819
            Take-off      0.92      0.90      0.91      1310
            Unknown      0.80      0.76      0.78       684

           accuracy                          0.87     10156
          macro avg      0.87      0.87      0.87     10156
       weighted avg      0.88      0.87      0.88     10156
```

Fig. 5. Classification report for the best model (LSTM)

The extract in Fig. 5 shows the classification report in terms of accuracy, precision, recall, and F1 score for the LSTM model. The extract also gives an account of the test instance distribution among distinct phase(s) of flight entries as evidenced in the support column. On the other hand, Fig. 6 gives a visual account of how the LSTM model distributes test instances in the form of a confusion matrix

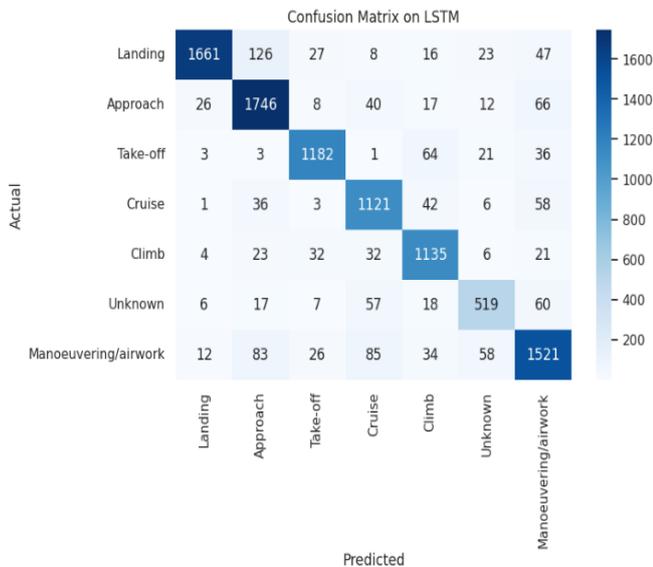

Fig. 6. Confusion Matrix

## V. CONCLUSION

In an era where aviation safety is paramount, the automation and precision of safety occurrence analysis take center stage. This research has showcased the immense potential of Natural Language Processing (NLP) and Deep Learning techniques in revolutionizing the classification of flight phases within safety occurrence reports. Leveraging a dataset of 50,778 reports from the Australian Transport Safety Bureau (ATSB), we employed advanced Deep Learning architectures, including Long Short-Term Memory (LSTM), Convolutional Neural Network (CNN), Bidirectional LSTM (BLSTM), and simple Recurrent Neural Network (sRNN), to infer critical flight phase information from unstructured text narratives.

Our findings are not only promising but transformative. The LSTM, BLSTM, and CNN models exhibited remarkable performance, as evidenced by high precision, recall, and F1-score values. This exceptional performance underscores the capacity of NLP and Deep Learning models to handle the complexity of aviation safety reports, facilitating precise flight phase classification.

The implications of this research extend beyond the academic realm. The integration of NLP and Deep Learning technologies into aviation safety analysis promises to streamline the identification of safety trends, enhance contextual awareness, and inform targeted safety measures. The ability to associate safety occurrences with specific flight phases allows regulatory authorities and aviation industry stakeholders to develop proactive and context-aware safety protocols. Furthermore, the scalability and automation of this approach empower aviation safety experts to handle a higher volume of reports efficiently. This ensures that critical safety information is not overlooked and that safety enhancements can be implemented promptly, contributing to a safer aviation environment.

Therefore, the findings presented in this research signify a transformative shift in aviation safety analysis. The synergy of NLP and Deep Learning technologies not only augments the efficiency of safety occurrence analysis but also elevates the precision and contextual understanding of aviation safety incidents. Future work in this field holds great promise, including further model refinement, the integration of multimodal data sources, real-time analysis capabilities, cross-dataset validation, improved model interpretability, and the development of human-in-the-loop systems. As we continue to explore these avenues, we are poised to strengthen aviation safety measures and contribute to the ongoing mission of safer skies for all.